\documentclass[sigconf]{acmart}
\usepackage{multirow}
\usepackage{subcaption}

\copyrightyear{2026}
\acmYear{2026}
\setcopyright{cc}
\setcctype{by}
\acmConference[CIKM '26]{Proceedings of the 35th ACM International Conference on Information and Knowledge Management}{November 07--11, 2026}{Rome, Italy}
\acmBooktitle{Proceedings of the 35th ACM International Conference on Information and Knowledge Management (CIKM '26), November 07--11, 2026, Rome, Italy}
\acmDOI{10.1145/3799682.3839912}
\acmISBN{979-8-4007-2539-5/2026/11}

\AtBeginDocument{%
  }

\begin{document}

\title[MINT]{MINT: Mitigating Structural Inconsistency via Attentive Message Passing for Financial Anomaly Detection}

\author{Yin-Ju Wu}
\affiliation{%
  \institution{National Taiwan Normal University}
  \city{Taipei}
  \country{Taiwan}}
\email{61247075s@ntnu.edu.tw}

\author{Bi-Cheng Yan}
\affiliation{%
  \institution{National Taiwan Normal University}
  \city{Taipei}
  \country{Taiwan}}
\email{80847001s@ntnu.edu.tw}

\author{Berlin Chen}
\affiliation{%
  \institution{National Taiwan Normal University}
  \city{Taipei}
  \country{Taiwan}}
\email{berlin@ntnu.edu.tw}


\begin{abstract}
Illicit account detection is a critical task for financial risk control systems, aiming to pinpoint malicious actors within transaction networks. Existing efforts typically recast transaction networks into directed multigraphs and leverage graph neural networks (GNNs) to detect anomalies via topological structures under the homophily assumption. However, directly grafting GNNs onto transaction graphs is faced with at least two pivotal challenges, i.e., structural inconsistency and label scarcity. The former arises from deceptive connections forged by malicious actors, while the latter stems from the lack of annotated data. In light of these, we introduce MINT~\footnote{Source code and models are publicly available at: https://github.com/innroowu/MINT.git}, a novel graph neural framework that \underline{m}itigates structural \underline{in}consistency via an attentive message-passing scheme on \underline{t}ransaction graphs. Specifically, MINT introduces a discrepancy-aware feature propagation, which tracks malicious actors by quantifying deviations between nodes across bi-directional transaction flows. A transaction salience module is proposed to mitigate structural inconsistency by adaptively regularizing the interaction strength between nodes. Furthermore, to alleviate label scarcity, MINT is optimized via Positive-Unlabeled (PU) learning, which exploits only a minority of illicit nodes coupled with a majority of unlabeled ones. Extensive experiments conducted on three financial transaction benchmarks, including two large-scale public and one in-house real-world datasets, validate the superior effectiveness of MINT over several cutting-edge baselines.
\end{abstract}



\maketitle

\section{Introduction}
Graph Anomaly Detection (GAD) has emerged as a pivotal task in transaction monitoring and financial risk control systems, dedicated to identifying atypical graph entities (nodes and edges) that significantly deviate from global structures or local affinities~\cite{SemiGAN, MAD}. Since large-scale financial transaction networks naturally manifest as directed multigraphs, extensive studies have leveraged GAD techniques to detect illicit accounts~\cite{gfp, amlinbitcoin, gnnreview, survey, LABRA}. By tracking transaction flows via iterative message passing, GAD approaches harness graph neural networks (GNNs) to effectively pinpoint malicious actors or fraudulent behaviors based on topological structures~\cite{gcn, gat, sage}.

\begin{figure}[t]
  \centering
  \includegraphics[width=\linewidth]{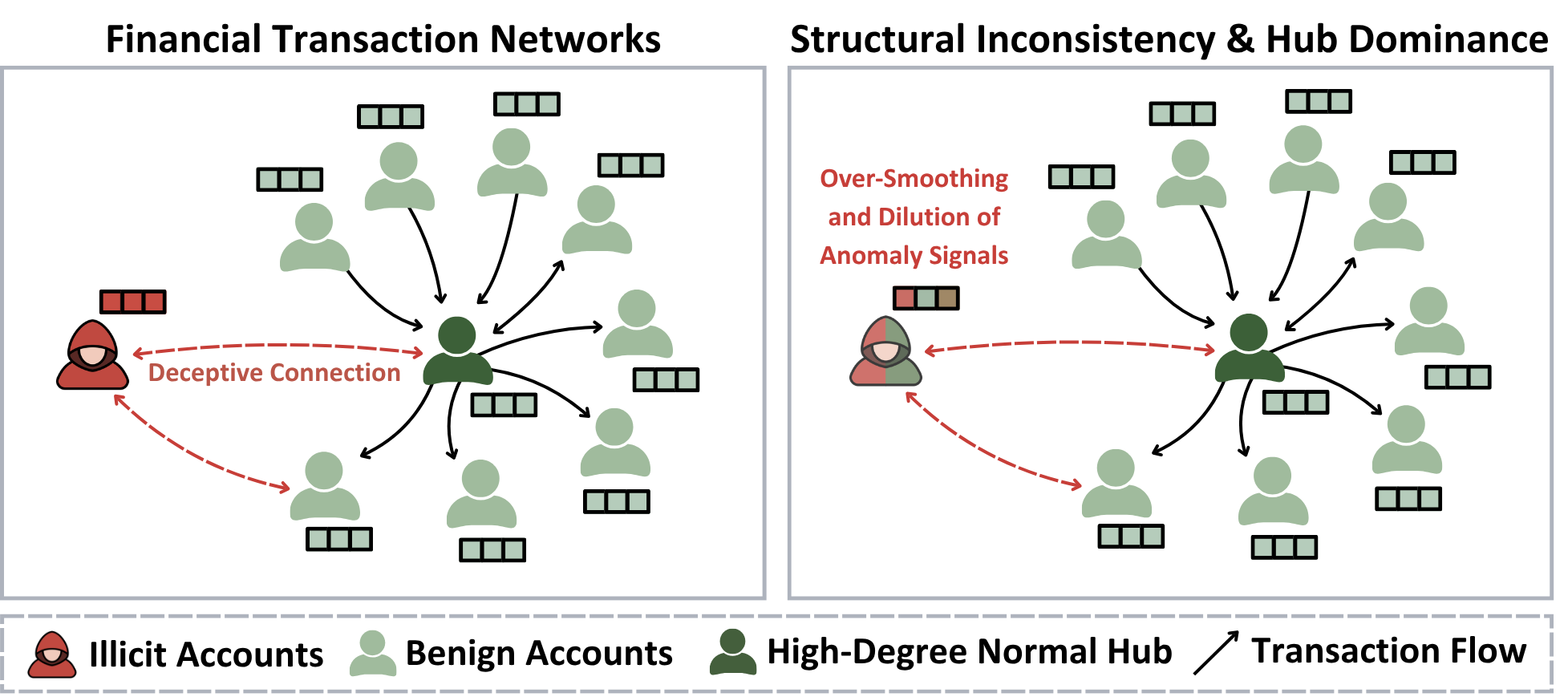}
  \caption{Illustration of structural inconsistency and transaction hub dominance in financial transaction networks.}
  \label{fig:illustration}
\end{figure}

Existing efforts have applied GNNs to detect financial anomalies, predominantly relying on the homophily assumption where connected nodes share similar attributes or local structures~\cite{Beyond_Homophily, Beyond_homophily_GNN}. However, this assumption is untenable in real-world transaction networks, where the neighborhood aggregation mechanism of GNNs inevitably dilutes the subtle signals of graph anomalies. Directly grafting GNN-based approaches onto illicit account detection poses two pivotal challenges, namely structural inconsistency and label scarcity. Specifically, as illustrated in Figure~\ref{fig:illustration}, the structural inconsistency arises from malicious actors forging deceptive connections with benign accounts, while high-degree nodes (i.e., transaction hubs) tend to dominate node representation learning. The latter stems from the fact that accurate anomaly annotation is highly labor-intensive and requires exhaustive auditing by domain experts, resulting in a scarcity of labeled nodes while leaving the vast majority unlabeled~\cite{GAD_fewLabels, Finance_review}.

Building on these observations, we propose MINT, a novel graph neural framework tailored for transaction graphs that effectively mitigates structural inconsistency via attentive message passing. Specifically, to capture anomaly signals over directed multigraphs, MINT introduces a discrepancy-aware message-passing scheme, which aims to iteratively quantify feature deviations between nodes across their incoming and outgoing neighborhoods. Furthermore, to mitigate structural inconsistency, MINT incorporates a structural edge reweighting strategy into the message-passing scheme to dynamically calibrate connection strengths based on transaction statistics. In contrast to traditional attentive message-passing which relies on feature similarity to evaluate node importance, our reweighting strategy modulates connection intensities based on two transactional factors, namely, the local transaction frequency and the global transaction rarity of a node. To mitigate label scarcity, MINT is optimized within a Positive-Unlabeled (PU) learning paradigm, only leveraging a minority of labeled illicit nodes coupled with a vast majority of unlabeled ones during the training phase~\cite{pu_imbalanced, puloss, LearnClassifiersfromPU}. Extensive experiments on three financial transaction benchmarks, including two large-scale public and one real-world proprietary datasets, confirm the superior effectiveness and robustness of MINT over various cutting-edge baselines. The main contributions of this work are summarized as follows:
\begin{itemize}
\item We identify structural inconsistency and extreme label scarcity as primary bottlenecks in financial anomaly detection, where deceptive connections and high-degree hubs jointly dilute rare anomalous signals.
\item We propose MINT, a framework integrating discrepancy-aware message passing and structural edge reweighting to dynamically calibrate connection strengths, optimized under a Positive-Unlabeled setting to overcome label scarcity.
\item Extensive experiments on public benchmarks and real-world financial dataset demonstrate the superior effectiveness and robustness of MINT against cutting-edge baselines.
\end{itemize}

\section{Preliminaries}
\textbf{Graph Construction.} We formalize the financial transaction network as a directed multigraph $\mathcal{G} = (\mathcal{V}, \mathcal{E}, \mathbf{X})$, where $\mathcal{V}$ represents the account set and each edge $e_{u,v} \in \mathcal{E}$ denotes a directed transaction from $u$ to $v$. The transactional behaviors of accounts are encoded within a feature matrix $\mathbf{X} \in \mathbb{R}^{|\mathcal{V}| \times d}$, mapping each node $v_i$ to a statistical profile vector $\mathbf{x}_i \in \mathbb{R}^d$ derived from interaction frequencies and monetary aggregates, which serves as the initial state $\mathbf{h}_i^{(0)} = \mathbf{x}_i$ for the $l$-th layer hidden representation $\mathbf{h}_i^{(l)}$ during message passing.

\textbf{Problem Formulation.} Financial anomaly detection on $\mathcal{G}$ is formulated as a PU node classification problem, partitioning the account set as $\mathcal{V} = \mathcal{P} \cup \mathcal{U}$ with $\mathcal{P} \cap \mathcal{U} = \emptyset$. Since $\mathcal{P}$ contains only a small fraction of verified illicit nodes and $\mathcal{U}$ represents an unknown mixture of benign and illicit accounts, the strict absence of verified negative labels necessitates learning a graph-based scoring function $f: (\mathcal{G}, \mathbf{X}) \rightarrow [0, 1]^{|\mathcal{V}|}$ to accurately estimate the illicit probability for each node in $\mathcal{U}$.

\section{Methodology}

\begin{figure*}[t]
  \centering
  \includegraphics[width=0.85\linewidth]{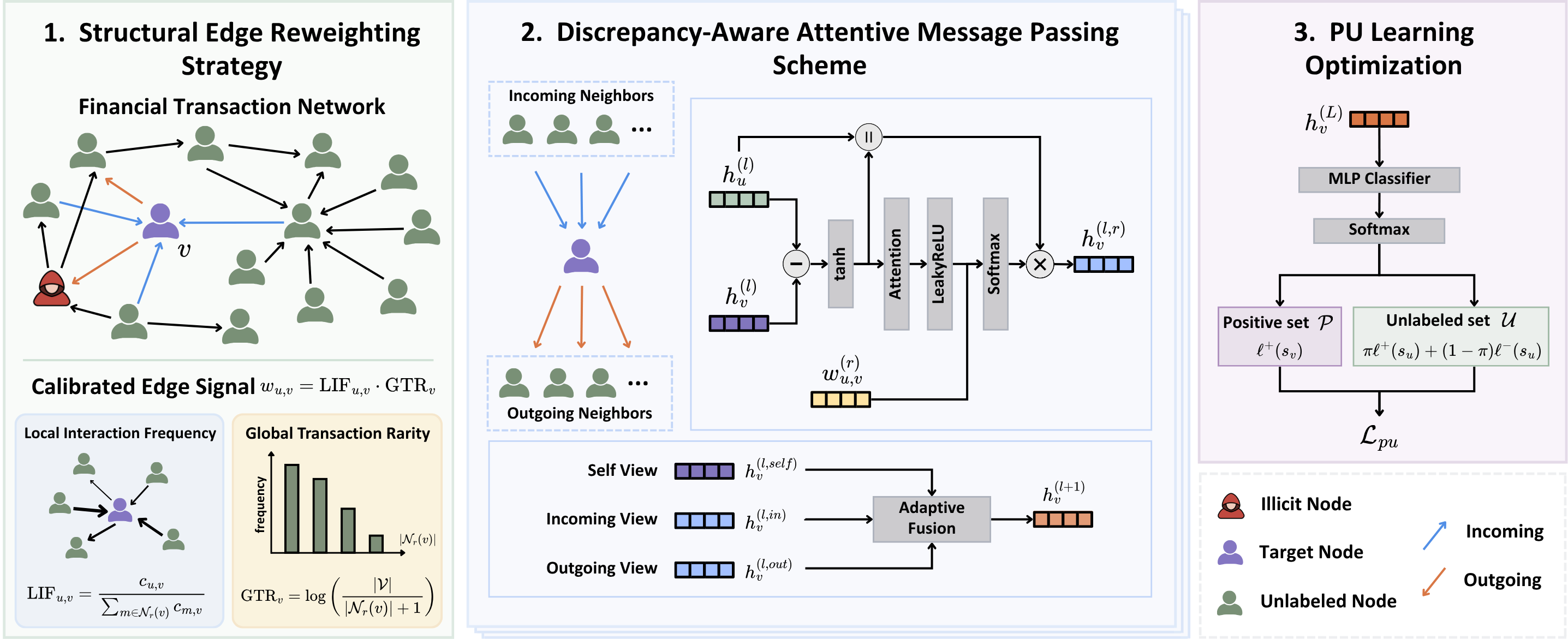}
  \caption{Overview of the proposed MINT framework.}
  \label{fig:framework}
\end{figure*}

In this section, we detail the proposed MINT framework as illustrated in Figure~\ref{fig:framework}. Our framework integrates a structural edge reweighting strategy and a discrepancy-aware attentive message-passing scheme. The entire framework is optimized with a PU learning objective that effectively stabilizes training under extreme label scarcity scenarios.

\subsection{Structural Edge Reweighting Strategy}
\label{sec:LIF_GTR}
In financial transaction graphs, representation learning is prone to be overwhelmed by high-degree nodes ( transaction hubs), which inevitably dilutes the signals of graph anomalies. To address this limitation, MINT introduces a structural edge reweighting strategy for feature propagation, dynamically calibrating the connection intensity of each transaction by factoring in two topological statistics, namely, the local interaction frequency of the node pair and the global transaction rarity of a node. Furthermore, to capture nuanced graph anomalies concealed within asymmetric directional transaction flows, the proposed calibration strategy is independently computed over incoming and outgoing edge relations, denoted by $r \in \{\mathrm{in}, \mathrm{out}\}$.

For a given relation $r$, the calibrated connection strength between nodes $u$ and $v$ is denoted as:
\begin{equation}
w_{u,v}^{(r)} = \mathrm{LIF}_{u,v}^{(r)} \cdot \mathrm{GTR}_{v}^{(r)},
\end{equation}
where the Local Interaction Frequency ($\mathrm{LIF}$) quantifies the relative transaction intensity from a source account $u$ to a target account $v$, and the Global Transaction Rarity ($\mathrm{GTR}$) measures the transactional activity of node $v$. Both mathematical terms are defined as follows:
\begin{equation}
\mathrm{LIF}_{u,v}^{(r)} = \frac{c_{u,v}^{(r)}}{\sum_{m \in \mathcal{N}_r(v)} c_{m,v}^{(r)}},
\end{equation}
\begin{equation}
\mathrm{GTR}_{v}^{(r)} = \log \left( \frac{|\mathcal{V}|}{|\mathcal{N}_r(v)| + 1} \right),
\end{equation}
where $c_{u,v}^{(r)}$ denotes the transaction count from node $u$ to node $v$ within the relation-specific neighborhood $\mathcal{N}_r(v)$, and $|\mathcal{V}|$ represents the total number of accounts in the graph. By leveraging the synergy between $\mathrm{LIF}_{u,v}^{(r)}$ and $\mathrm{GTR}_{v}^{(r)}$, MINT effectively captures local connectivity between nodes while penalizing the oversmoothing effect induced by transaction hubs.

\subsection{Discrepancy-aware Attentive Message-passing Scheme}
To mitigate the structural inconsistency caused by deceptive connections, MINT departs from conventional feature oversmoothing by explicitly quantifying representational deviations across directed monetary flows. Specifically, at the $l$-th message-passing layer, the representational deviation between a target account $v$ and its neighbor $u$ under relation $r$ is measured as:
\begin{equation}
\mathbf{d}_{v,u}^{(l,r)} = \tanh(\mathbf{h}_v^{(l)} - \mathbf{h}_u^{(l,r)}),
\end{equation}
where $\mathbf{h}_v^{(l)}$ and $\mathbf{h}_u^{(l,r)}$ denote the node representations of accounts $v$ and $u$, respectively. Subsequently, to obtain the discrepancy feature $\mathbf{h}_{v}^{(l,r)}$ for node $v$, MINT incorporates the calibrated edge signal $w^{(r)}_{u,v}$ (derived in Section 3.1) into the attentive message-passing scheme as a structural attention coefficient, aggregating directional messages from the neighborhood $u \in \mathcal{N}_r(v)$:
\begin{equation}
\alpha_{u,v}^{(l,r)} = \mathrm{softmax} \left( \mathrm{LeakyReLU}(\mathbf{a}^\top \mathbf{d}_{v,u}^{(l,r)}) + \lambda_e w_{u,v}^{(r)} \right);
\end{equation}
\begin{equation}
\mathbf{h}_{v}^{(l,r)} = \mathbf{W}_c \sum_{u \in \mathcal{N}_r(v)} \alpha_{u,v}^{(l,r)} [\mathbf{d}_{v,u}^{(l,r)} \parallel \mathbf{h}_u^{(l,r)}],
\end{equation}
where $\mathbf{a}$ denotes a learnable weight vector, $\lambda_e$ is a learnable scaling factor for adjusting the relative impact of node discrepancies and transactional statistics, and $\mathbf{W}_c$ acts as a transformation matrix.

Finally, the node representation for account $v$ is updated by aggregating its intrinsic representation $\mathbf{h}_{v}^{(l)}$ in conjunction with the relation-specific discrepancy features (i.e., $\mathbf{h}_{v}^{(l,\mathrm{in})}$ and $\mathbf{h}_{v}^{(l,\mathrm{out})}$) via a simple attention mechanism. Formally, the updated representation $\mathbf{h}_v^{(l+1)}$ is formulated as:
\begin{equation}
\beta^{(k)} = \frac{\exp(\mathbf{w}_k^\top \mathbf{h}_{v}^{(l,k)})}{\sum_{k \in \{\mathrm{self}, \mathrm{in}, \mathrm{out}\}} \exp(\mathbf{w}_k^\top \mathbf{h}_{v}^{(l,k)})};
\end{equation}
\begin{equation}
\mathbf{h}_v^{(l+1)} = \mathbf{W}_f \sum_{k \in \{\mathrm{self}, \mathrm{in}, \mathrm{out}\}} \beta^{(k)} \mathbf{h}_{v}^{(l,k)},
\end{equation}
where $\mathbf{W}_f$ denotes a learnable transformation matrix, $\mathbf{w}_k$ represents the attention weight vector for each $k \in \{\mathrm{self}, \mathrm{in}, \mathrm{out}\}$, and $\mathbf{h}_{v}^{(l,\mathrm{self})}$ is set as $\mathbf{h}_{v}^{(l)}$.

\subsection{Positive-Unlabeled Learning Optimization}
To derive the illicit probability $s_v$, we project the final embedding $\mathbf{h}_v^{(L)}$ through an MLP classifier equipped with softmax normalization. To robustly accommodate extreme label sparsity, we optimize the model under the established PU setting. Rather than naively treating the unverified mixture $\mathcal{U}$ as strict negatives, we mitigate false-negative bias by approximating the expected risk via a prior probability $\pi$, which serves as the expected risk representing the estimated illicit ratio. Defining $\ell^+$ and $\ell^-$ as the standard positive and negative cross-entropy losses, the objective is formulated as:
\begin{equation}
\mathcal{L}_{\mathrm{PU}} = \frac{1}{|\mathcal{P}|} \sum_{v \in \mathcal{P}} \ell^+(s_v) + \frac{1}{|\mathcal{U}|} \sum_{v \in \mathcal{U}} \left[ \pi \ell^+(s_v) + (1-\pi) \ell^-(s_v) \right]
\end{equation}
By circumventing the forced assignment of negative labels onto unverified accounts, this prior-weighted optimization empowers MINT to accurately identify malicious actors under strictly limited supervision.

\section{Experiments}

\subsection{Experimental Settings}

\begin{table}[t]
  \caption{Statistics of the financial transaction datasets.}
  \label{tab:datasets}
  \begin{tabular}{lrrr}
    \toprule
    Dataset & \# Nodes & \# Edges & \# Anomaly (Rate) \\
    \midrule
    RealFT & 1,800,106 & 4,435,890 & 1,004 (0.056\%) \\
    Elliptic & 822,942 & 2,868,964 & 4,280 (0.520\%) \\
    AMLSim & 10,000 & 1,323,234 & 506 (5.060\%) \\
  \bottomrule
\end{tabular}
\end{table}

\textbf{Datasets and Metrics.} To emulate real-world supervision scenarios, we evaluate MINT within a PU learning framework across three financial transaction networks (detailed in Table~\ref{tab:datasets}): a proprietary bank transfer dataset (RealFT), a cryptocurrency network (Elliptic)~\cite{ellipticPlusPlus}, and a synthetic anti-money laundering dataset (AMLSim)~\cite{amlsim}. RealFT consists of de-identified bank transfer records provided by a financial institution and covers a three-month observation period. The dataset contains tabular transaction records describing fund transfers between accounts, while illicit accounts are identified through manual investigation by domain experts. Given the extreme class imbalance, we prioritize AUPRC to robustly assess illicit ranking capabilities~\cite{PRC_imbalanced}, supplemented by AUROC and F1-score~\cite{ROC, PRCROC}, with all reported metrics averaged over five independent trials to guarantee statistical stability.

\begin{table*}[t]
\centering
\caption{Performance comparison across three datasets. Results are reported as mean $\pm$ standard deviation.}
\scriptsize
\label{tab:main_results}
\setlength{\tabcolsep}{4pt}
\renewcommand{\arraystretch}{0.95}
\resizebox{\textwidth}{!}{
\begin{tabular}{lccccccccc}
\toprule
& \multicolumn{3}{c}{RealFT}
& \multicolumn{3}{c}{AMLSim}
& \multicolumn{3}{c}{Elliptic} \\
\cmidrule(lr){2-4}
\cmidrule(lr){5-7}
\cmidrule(lr){8-10}
Method
& F1 & AUPRC & AUROC
& F1 & AUPRC & AUROC
& F1 & AUPRC & AUROC \\
\midrule

GCN
& 27.62$\pm$2.06 & 22.30$\pm$0.96 & 78.33$\pm$0.46
& 36.23$\pm$1.58 & 37.38$\pm$1.30 & 69.58$\pm$1.07
& 18.25$\pm$1.25 & 10.82$\pm$0.59 & 82.83$\pm$0.68 \\

GAT
& 16.79$\pm$3.22 & 12.69$\pm$2.01 & 70.72$\pm$4.12
& 34.99$\pm$4.57 & 25.97$\pm$3.50 & 64.38$\pm$6.42
& 11.41$\pm$4.99 & 8.64$\pm$1.52 & 80.69$\pm$1.99 \\

GraphSAGE
& 49.41$\pm$2.04 & 51.03$\pm$1.25 & 86.10$\pm$0.78
& 80.81$\pm$1.62 & 87.82$\pm$0.55 & 93.68$\pm$0.29
& 35.76$\pm$1.76 & 30.69$\pm$1.21 & 91.84$\pm$0.15 \\
\midrule

SEC-GFD
& 37.00$\pm$1.12 & 36.86$\pm$1.73 & 83.22$\pm$0.82
& 81.48$\pm$0.72 & 87.56$\pm$0.53 & 92.59$\pm$0.36
& 38.86$\pm$0.35 & 34.81$\pm$1.07 & 91.99$\pm$0.19 \\

FRAUDRE
& 43.41$\pm$2.95 & 43.31$\pm$2.38 & 85.92$\pm$0.81
& 83.27$\pm$0.57 & 87.63$\pm$0.60 & 92.63$\pm$0.62
& 35.80$\pm$1.08 & 32.63$\pm$1.08 & 91.80$\pm$0.22 \\

ConsisGAD
& 47.19$\pm$3.30 & 51.13$\pm$2.95 & 86.00$\pm$1.20
& 82.06$\pm$0.88 & 86.75$\pm$0.42 & 91.81$\pm$0.23
& 29.80$\pm$0.54 & 22.07$\pm$0.87 & 89.26$\pm$0.21 \\

DIAM
& 58.15$\pm$5.61 & 61.76$\pm$6.74 & 92.14$\pm$1.48
& 79.60$\pm$0.67 & 84.25$\pm$2.32 & 92.34$\pm$0.86
& 40.65$\pm$1.17 & 38.73$\pm$0.60 & 92.14$\pm$0.80 \\
\midrule

MINT
& 70.13$\pm$2.02 & 76.94$\pm$1.25 & 95.37$\pm$0.36
& 80.34$\pm$1.14 & 87.03$\pm$2.27 & 93.29$\pm$0.34
& 56.90$\pm$2.81 & 55.30$\pm$4.39 & 97.14$\pm$0.46 \\

\bottomrule
\end{tabular}
}
\end{table*}

\textbf{Comparative Methods.} To evaluate MINT, we compare it against seven baselines across two paradigms. For general-purpose GNNs, we select GCN~\cite{gcn}, GAT~\cite{gat}, and  GraphSAGE~\cite{sage}, representing foundational approaches that capture structural dependencies via isotropic neighborhood aggregation, attention mechanisms, and localized sampling, respectively. To assess specialized GAD capabilities, we include advanced architectures tailored for complex network challenges. FRAUDRE~\cite{FRAUDRE} handles graph inconsistency and label imbalance via fraud-aware convolution and an imbalance-oriented classifier. ConsisGAD~\cite{ConsisGAD} improves limited-supervision anomaly detection using a homophily-aware backbone with learnable consistency augmentation. SEC-GFD~\cite{secgfd} addresses heterophily through spectrum-enhanced filtering and local environmental constraints. DIAM~\cite{diam} captures multi-scale behavioral discrepancies between connected nodes in cryptocurrency networks.

\textbf{Implementation Details.} Simulating real-world label scarcity, RealFT uses full annotations, while AMLSim and Elliptic retain only 30\% of verified illicit nodes as known positives, with 10\% held out for threshold and PU-prior tuning. MINT is implemented in PyTorch with 4 layers, dropout 0.3, hidden dimensions of 40/56/30 for RealFT/AMLSim/Elliptic, and trained for 300 epochs using Adam with a learning rate of 0.005 and weight decay of $5\times10^{-4}$ on NVIDIA TITAN RTX GPUs.

\subsection{Comparative Results}
As presented in Table~\ref{tab:main_results}, MINT exhibits a substantial performance advantage on the real-world RealFT and Elliptic datasets. These results achieve substantial absolute AUPRC gains of 15.18\% and 16.57\% over DIAM, the strongest specialized baseline, empirically validating the effectiveness of our proposed framework under extreme label scarcity. Conversely, foundational GNN models inherently underperform in such contexts. This limitation confirms that strict homophily-based neighborhood aggregation is highly vulnerable to hub-induced over-smoothing, which blurs subtle anomalous signals into dominant benign hubs, and remains incapable of capturing asymmetric monetary flows. While specialized models like FRAUDRE and DIAM demonstrate robust performance on real-world datasets, GraphSAGE secures a marginal lead on the synthetic AMLSim dataset via localized feature sampling. Nevertheless, MINT remains highly competitive even in this scenario, reaffirming its overall robustness and decisive superiority for illicit account detection across diverse financial networks.

\subsection{Ablation Study}

\begin{table}[t]
  \caption{Ablation study of MINT on the RealFT dataset.}
  \label{tab:ablation}
  \scriptsize
  \centering
  \resizebox{\columnwidth}{!}{
  \begin{tabular}{llccc}
    \toprule
    Module & Variant & F1 & AUPRC & AUROC \\
    \midrule
    MINT & -- & \textbf{70.13} & \textbf{76.94} & \textbf{95.37} \\
    \midrule
    \multirow{2}{*}{Direction}
    & w/o Outgoing & 38.14 & 36.90 & 88.02  \\
    & w/o Incoming & 44.68 & 49.43 & 87.56  \\
    \midrule
    \multirow{3}{*}{Edge Signal}
    & w/o LIF & 67.85 & 78.86 & 95.64  \\
    & w/o GTR & 63.04 & 78.10 & 95.59  \\
    & w/o LIF+GTR  & 68.42 & 74.15 & 93.85  \\
    \bottomrule
  \end{tabular}
  }
\end{table}

We conduct an ablation study on the RealFT dataset to assess the contribution of each major component in MINT. As shown in Table~\ref{tab:ablation}, eliminating either the incoming or outgoing aggregation paths triggers severe AUPRC degradation, empirically substantiating the necessity of modeling asymmetric monetary flows. Furthermore, ablating the structural edge reweighting strategy inherently degrades overarching AUPRC and AUROC, while removing its individual components induces decline in F1 scores. These results confirm that jointly calibrating local interaction frequency and global transaction rarity is indispensable for robust illicit account detection.

\subsection{Visualization Analysis}

\begin{figure}[t]
  \centering
  
  \begin{subfigure}{0.48\columnwidth}
    \centering
    \includegraphics[width=\linewidth]{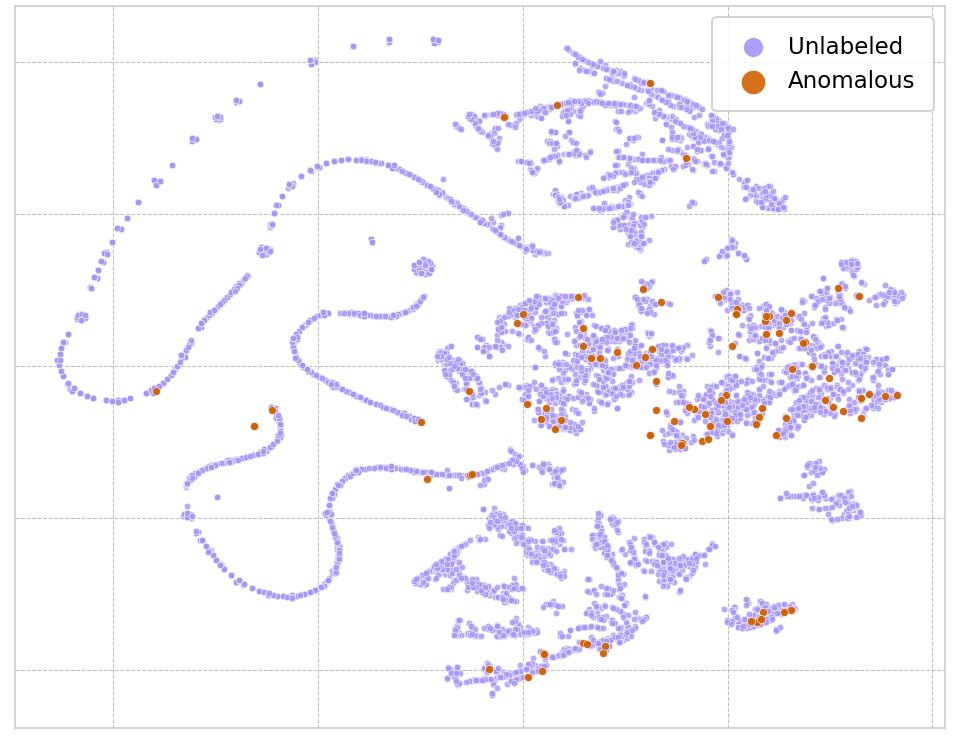}
    \caption{RealFT (Raw)}
  \end{subfigure}
  \hfill
  \begin{subfigure}{0.48\columnwidth}
    \centering
    \includegraphics[width=\linewidth]{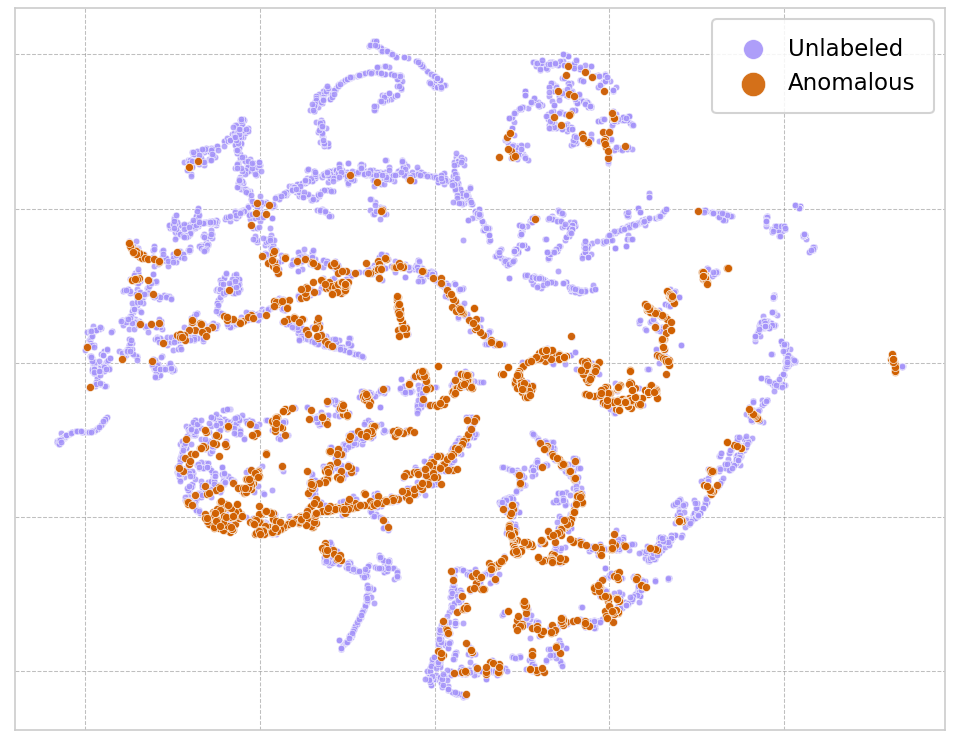}
    \caption{Elliptic (Raw)}
  \end{subfigure}

  \vspace{2mm} 

  \begin{subfigure}{0.48\columnwidth}
    \centering
    \includegraphics[width=\linewidth]{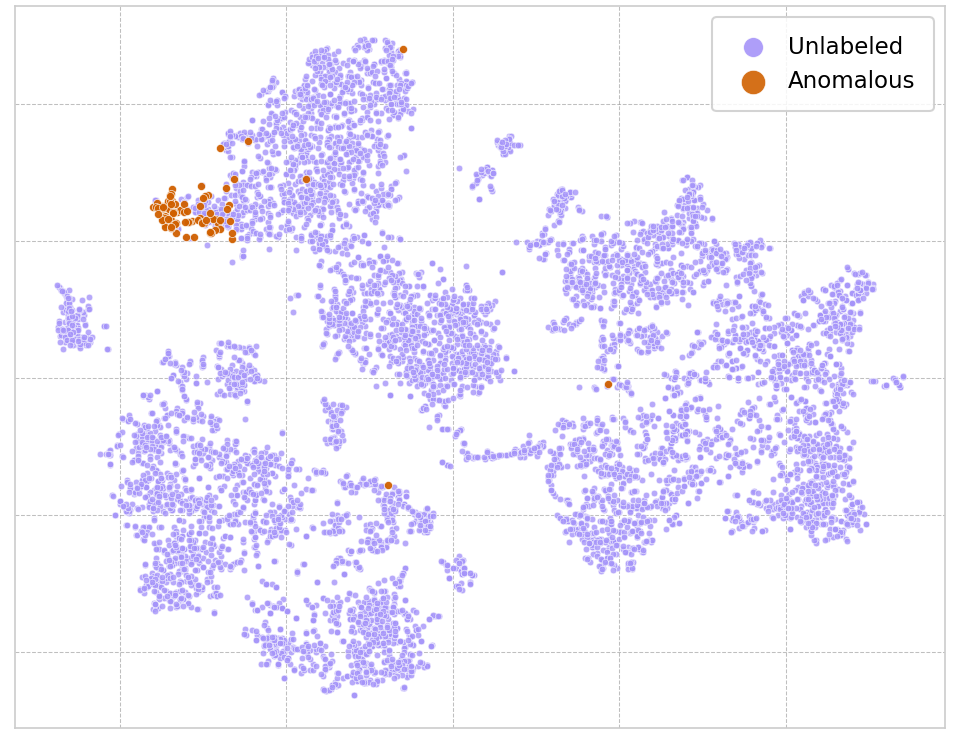}
    \caption{RealFT (MINT)}
  \end{subfigure}
  \hfill
  \begin{subfigure}{0.48\columnwidth}
    \centering
    \includegraphics[width=\linewidth]{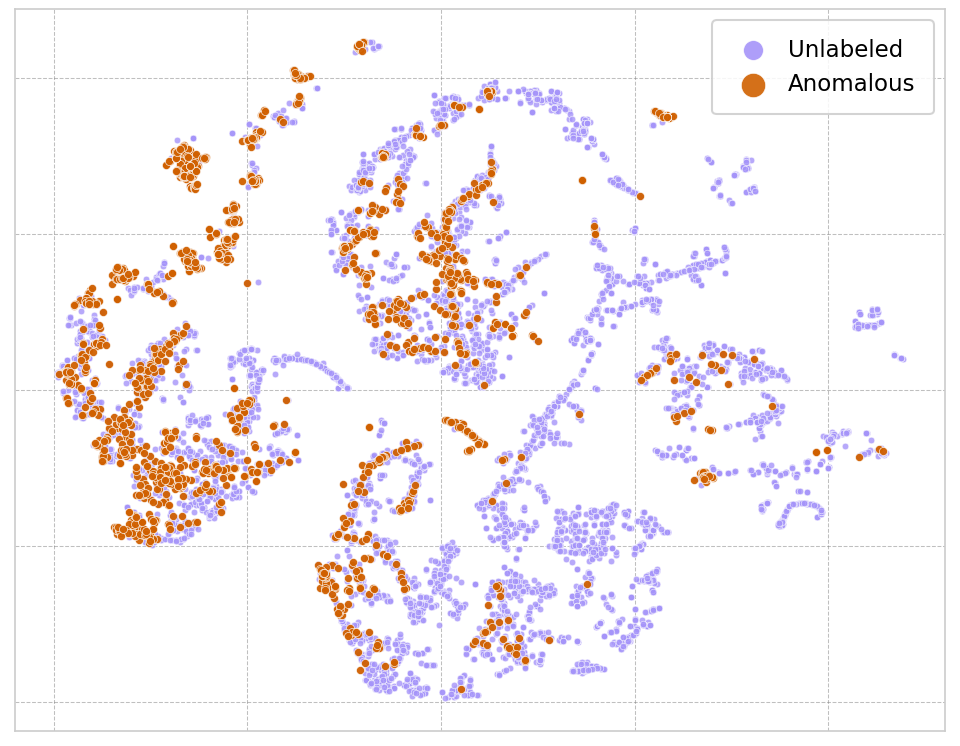}
    \caption{Elliptic (MINT)}
  \end{subfigure}

  \caption{t-SNE visualization of node representations learned by MINT on RealFT and Elliptic.}
  \label{fig:tsne}
\end{figure}

Figure~\ref{fig:tsne} compares initial features and MINT embeddings via t-SNE. Initially, illicit and unlabeled nodes are heavily entangled, reflecting severe structural inconsistency in transaction graphs. Conversely, MINT separates these illicit nodes into more compact and distinguishable regions. This qualitative evaluation provides intuitive evidence that MINT effectively mitigates structural inconsistency and hub-induced over-smoothing, yielding discriminative representations for rare financial anomalies.

\section{Conclusion}
In this paper, we introduce MINT, a graph neural framework resolving structural inconsistency and extreme label scarcity in financial anomaly detection. We integrate a structural edge reweighting strategy and a discrepancy-aware attentive message passing scheme to neutralize the structural inconsistency caused by deceptive connections and transaction hub dominance, while capturing feature deviations across neighborhoods. Extensive experiments across diverse large-scale transaction benchmarks demonstrate that MINT, optimized under a Positive-Unlabeled learning framework, achieves superior effectiveness and robustness.

\section*{Acknowledgments}
This work was supported in part by E.SUN Commercial Bank, Ltd., Taiwan, under Grant Number 202508-NTU-03. The authors would like to express their gratitude to the collaborators at E.SUN Commercial Bank, Yi-Ren Yeh, Fu-Chang Sun, and Wei Luok Ngu, for their valuable support and insights. Any findings and implications in the paper do not necessarily reflect those of the sponsor.

\section*{GenAI Usage Disclosure}
We declare that all content within this paper was independently conducted by the authors. GenAI tools were exclusively used for language polishing, improving grammatical correctness, and refining the flow of the text.

\bibliographystyle{ACM-Reference-Format}
\bibliography{reference}

\end{document}